\documentclass[letterpaper, 10 pt, conference]{ieeeconf}  

\IEEEoverridecommandlockouts                              
\title{\LARGE \bf
Distributed Block Coordinate Moving Horizon Estimation for 2D Visual-Inertial-Odometry SLAM
}

\usepackage[autostyle]{csquotes}
\usepackage{amsmath,verbatim,amssymb,amsfonts,amscd, graphicx}
\usepackage{mathabx}
\usepackage{breqn}
\usepackage{float}
\usepackage{graphics}
\usepackage{stmaryrd} 
\usepackage{textcomp}
\usepackage{setspace}
\usepackage{algorithm}
\usepackage{algpseudocode}
\usepackage{optidef}
\usepackage{caption, subcaption}
\usepackage{bbm}

\author{Emilien Flayac$^{1}$,  Iman Shames $^{2}$, 
    \thanks{$^{1}$ Emilien Flayac (emilien.flayac@isae.fr) (girish.nair@unimelb.edu.au) is with the Department of Complex Engineering System, ISAE-Supaero, Toulouse, France}%
    \thanks{$^{2}$Iman Shames (iman.shames@anu.edu.au) is with the  School of Engineering, The Australian National University, Canberra, Australia.}%
    \thanks{This work received funding from the Australian Government, via grant AUSMURIB000001 associated with ONR MURI grant N00014-19-1-2571}
}

\newtheorem{remark}{Remark}[section]

\begin{document}

\allowdisplaybreaks

\maketitle
\thispagestyle{empty}
\pagestyle{empty}

\begin{abstract}
This paper presents a Visual Inertial Odometry Landmark-based Simultaneous Localisation and Mapping algorithm based on a distributed block coordinate nonlinear Moving Horizon Estimation scheme. The main advantage of the proposed method is that the updates on the position of the landmarks are based on a Bundle Adjustment technique that can be parallelised over the landmarks. The performance of the method is demonstrated in simulations in different environments and with different types of robot trajectory. Circular and wiggling patterns in the trajectory lead to better estimation performance than straight ones, confirming what is expected from recent nonlinear observability theory.
\end{abstract}

\section{Introduction}

Visual Inertial and Visual Inertial Odometry Simultaneous Localisation and Mapping (VI-SLAM and VIO-SLAM) is the problem of localising a robot in a unknown environment while building a map of it using only visual, inertial and wheel odometry measurements.  VI-SLAM has gained a lot of attention in the recent decades due to the low cost and low energy consumption of cameras \cite{fuentes-pacheco_visual_2015} and the generalisation of Inertial Measurement Unit (IMU). SLAM problems are usually tackled using either filtering techniques or optimisation-based techniques. Typical filtering techniques include Extended Kalman or Information Filters (EKF-EIF) and Particle Filters (PF) \cite{thrun_probabilistic_2002}. EKFs are cheap and simple to implement but suffer from consistency issues due to successive linearisations and from bad scalability with respect to the number of landmarks considered in the environment \cite{strasdat_visual_2012,fuentes-pacheco_visual_2015,jia_survey_2019}. PFs are generally more precise and consistent than EKFs but substantially more computationally costly. Optimisation-based localisation and mapping techniques have recently proved to perform better than filtered-based methods for medium to large problems and at a reduced cost thanks to sparsification techniques \cite{huang_critique_2016, huang_review_2019}. However, proven techniques like Pose SLAM or Graph SLAM are mostly operated offline, \cite{thrun_graph_2006, ila_information-based_2010, valencia_mapping_2018}, while Bundle Adjustment has mainly been applied to purely visual settings, \cite{taketomi_visual_2017,bustos_visual_2019}. These methods are usually contained in the broader framework of Full Information Estimation (FIE) and Moving Horizon Estimation (MHE) framework, where the trajectory of a system is recovered by minimising the output error generated by the actual and predicted measurements, under a dynamical constraint. MHE is a simplified version of FIE where the optimisation is only performed on a sliding time window instead of starting from the initial time. Several VI-SLAM algorithms based on this idea have been designed, \cite{lupton_visual-inertial-aided_2012,bebis_monocular_2016,mur-artal_visual-inertial_2017,liu_ice-ba_2018,karrer_cvi-slamcollaborative_2018,kishimoto_moving_2019,heo_ekf-based_2019,campos_orb-slam3_2021}. However, the structure of the resulting optimisation problem does not seem to have been exploited yet. For example, in order to solve an MHE problem where the variables are the state of the system and fixed independent landmarks, one could iteratively fix the state variables and solve for the landmark variables and vice versa. This technique is called Block Coordinate Descent (BCD), see  Chapter 2 of \cite{bertsekas_nonlinear_1997}.  It has been applied to the Visual SLAM for example in \cite{strasdat_visual_2012},  PTAM SLAM \cite{klein_parallel_2007} and ORB-SLAM \cite{campos_orb-slam3_2021} sometimes under the denomination of \emph{motion-only} problem for state trajectory estimation or \emph{structure only} problem for landmark estimation. In these setup, the structure-only problem is typically high dimensional and become computationally costly.   In this paper, we propose a distributed BCD  method for Moving Horizon Estimation applied to landmark-based VI-O SLAM that allows one to parallelise the computations of landmark estimates. 

Many works in the field of robotics have showed that persistently excited path including circular ones are common sufficient conditions for good estimation using bearing measurements, \cite{deghat_target_2010, deghat_localization_2014, de_marco_position_2020, fossen_observers_2017, johansen_globally_nodate}. Thus, the performance and robustness  to noise of the proposed method depend on the trajectory of the robot. It is then demonstrated through simulations in several scenarios with different levels of excitations: a circular path in a circular corridor, a straight path in a straight corridor, a 'snaking' path in a straight corridor.  The rest of the paper is organised as follows: Section \ref{sec:dyn_meas} describes the dynamical and measurements models considered, Section \ref{sec:batch_MHE} presents a batch version of the MHE problem of interest, Section \ref{sec:bc_MHE} presents its block coordinated version, Section \ref{sec:algo} summarises the estimation algorithm and Section \ref{sec:simulations} gives simulation results.

\section{Dynamical and measurement models}\label{sec:dyn_meas} 
\subsection{Differential drive model}
We consider a mobile robot represented by a 2D position, $x=(x_1,x_2)\in \mathbb{R}^2$ and an orientation $\theta \in \mathbb{R}$. We assume it follows the differential drive dynamics such that:  
\begin{align}
       \dot{x}_1&=\frac{\omega_r R_r+\omega_l R_l}{2}\cos(\theta),\notag \\
       \dot{x}_2&=\frac{\omega_rR_r+\omega_lR_l}{2}\sin(\theta),\label{eq:diff_drive}\\
       \dot{\theta}&=\frac{\omega_rR_r-\omega_lR_l}{D}.\notag
   \end{align}
where:
   \begin{itemize}
        \item $R_r>0$, $R_l>0$ and $D>0$ are respectively the radius of the right wheels, the radius of the left wheels and the distance between the two sets of wheels.   
        \item $\omega_{ref}=(\omega_r,\omega_l): \mathbb{R}^+ \rightarrow \mathbb{R}^2$ represent the angular velocities of the right and left wheels. 
   \end{itemize}

By setting $z=(x,\theta)$, we can sum up \eqref{eq:diff_drive} as follows:
\begin{align}
    \dot{z}=f(z,\omega_{ref}),\label{eq:diff_drive_short}
\end{align}
   where $f:\mathbb{R}^3 \times \mathbb{R}^2\rightarrow \mathbb{R}^3$.
  
  Let $z_0 \in \mathbb{R}^{3}$ be a fixed initial condition and $t_0\geq0$ be the reference initial time. In the sequel, for $t\geq t_0$, $z(t)=(x(t),\theta(t))$ represents the solution of \eqref{eq:diff_drive_short} at time $t$ starting from $z_0$ with initial time $t_0$ and input $\omega_{ref}$.  
  
\subsection{Discretization scheme}
 
  In the sequel, we consider measurements obtained at discrete points in time with two different sampling rates.  With this in mind, let $\Delta_{odo}>0$ and $\Delta_{vis}>0$, be respectively the discretization step of odometry and visual measurements. We assume in the rests of the paper that $m=\frac{\Delta_{vis}}{\Delta_{odo}}\in \mathbb{N}^*$. 
 Thus, for $k\geq 0$ and $0\leq i\leq m$, we define $t_{k,i}$ such that:
  \begin{align}
      t_{k,i}=t_0+k\Delta_{vis}+i\Delta_{odo}. 
  \end{align}
  Note that  one has $t_{k,N}=t_{k+1,0}$ for any $k\geq0$. When no ambiguity is possible, we will denote $t_{k,0}$ by $t_k$ for any $k\geq0$.

   \subsection{Odometry and Inertial measurements }
  We  assume that  one does not have access to $\omega_{ref}$ but only to a noisy discretised version through odometry measurements. It is denoted by $\omega_{odo}$ and reads:
  \begin{align}
       \omega_{odo}(t)=\sum_{k=0}^{+\infty}\sum_{i=0}^{m} {\omega}_{k,i}\mathbf{1}_{\{t_{k,i}\leq t < t_{k,i+1} \}},\label{eq:odo_meas}
  \end{align}
  where $\mathbf{1}$ denotes the indicator function, ${\omega}_{k,i}=\omega_{ref}(t_{k,i})+d_{k,i}^{odo}$ and $(d_{k,i}^{odo})_{k\geq0, 0\leq i \leq m-1}$ is an i.i.d. sequence of centered gaussian perturbations with covariance $Q^{odo}$. Thus, for a sequence ${\omega}_{k,0:i}$ ,$0\leq i \leq m$, one can write the discretized dynamics between $t_k$ and $t_{k,i}$ as follows:
  \begin{align*}
      z(t_{k,i})=f_{dis,i}( z(t_{k}),{\omega}_{k,0:i})
  \end{align*}
  for some function $f_{dis,i}$
  
  We also assume that for any $k\geq0$, inertial measurements are processed and give information on the displacement of $z$ between $t_{k,0}$ and $t_{k+1,0}$ denoted by $u_k$ and defined as follows:
  \begin{align}
      u_k=z(t_{k+1,0})-z(t_{k,0})+d_k^{in}.\label{eq:inertial_meas_1}
  \end{align}
  where $d_k^{in}\in \mathbb{R}^3$ is a Gaussian perturbation of covariance $Q_{in}$ representing the error caused by the integration of inertial measurements.
  From \eqref{eq:inertial_meas_1} and $t\geq t_0$, one can define $u(t)$ similarly to \eqref{eq:odo_meas}:
  \begin{align}
      u(t)=\sum_{k=0}^{+\infty}{u}_k\mathbf{1}_{\{t_{k,0}\leq t < t_{k+1,0} \}},
  \end{align}

   \subsection{Landmark-based bearing measurement model}
  For $J\in \mathbb{N}^*$, let $\ell=\left(\ell^{(j)}\right)_{1\leq j\leq J} \in (\mathbb{R}^2)^J$ be a collection of $J$ landmarks represented by a 2D position. We assume, that for $1\leq j\leq J$ and $k\geq0$, if landmark $j$ is seen by the robot at time $t_k$,  a measurement of the direction between the robot and landmark $j$ in the body frame of the robot is obtained from visual information. It is denoted by $ y^{(j)}_k$ and defined formally as follows:
  \begin{align}
      y^{(j)}_k=a^{(j)}_k\left( R(-\theta(t_k))\frac{\ell^{(j)}-x(t_k)}{\Vert \ell^{(j)}-x(t_k)\Vert }+v_k^{(j)}\right),\label{eq:bearing}
  \end{align}
where $\Vert \cdot \Vert$ denotes the Euclidean norm, $ R(\theta)=\begin{bmatrix}
    \cos(\theta)&-\sin(\theta)\\\sin(\theta)&\cos(\theta)
\end{bmatrix}$, $v_k^{(j)} \in \mathbb{R}^2$ is an unknown perturbation representing the measurement noise, and $a^{(j)}_k$ is a data association parameter encoding the fact that landmark $j$ is seen at time $k$. Indeed, $a^{(j)}_k=1$ when landmark $j$ is seen at time $t_k$ and $a^{(j)}_k=0$ otherwise. We set $y_k=\left(y_k^{(j)}\right)_{1\leq j\leq J}$ and $v_k=\left(v_k^{(j)}\right)_{1\leq j\leq J}$ so that \eqref{eq:bearing} can be written as follows:
\begin{align}
    y_k=A_k(h(z(t_k),\ell)+v_k),\label{eq:visual_meas}
\end{align}
  where for any $k\geq0$, $h=\begin{bmatrix}h^{(1)}\\ \vdots\\h^{(J)} \end{bmatrix}$ with $h_k^{(j)}:\mathbb{R}^3\times \mathbb{R}^{2}\rightarrow \mathbb{R}^{2}$ for any $1\leq j\leq J$, and $A_k$ is the diagonal matrix of appropriate size with repetitions of binary numbers $a^{(j)}_k$ on its diagonal. Beside, we consider the sensor-centric view where the initial state is assumed known and can be considered as the origin of the robot frame. 
  
The goal of the following is to estimate the state of \eqref{eq:diff_drive_short} at time $t_{k,0}$ for any $k\geq0$, $z(t_{k,0}),$ and the position of the collection of landmarks, $\ell$ knowing the initial state $z_0$ and time $t_0$ . In particular, one is not interested in estimating $z$ at the times $t_{k,i}$ for $1\leq i\leq N-1$. 
 \section{Batch Moving horizon estimation for bearing-only SLAM}\label{sec:batch_MHE}
 \subsection{Discretized formulation of MHE}
Fix $N\in \mathbb{N}^*$ and let $T=N\Delta_{vis}$ be a time horizon.  In the sequel, for any $n\geq1$, we denote by $\mathbb{S}_n^{++}$ the set of positive definite $n\times n$ matrices.  Besides, for any   $S\in \mathbb{S}_n^{++}$, $\Vert\cdot \Vert_{S}$ denotes the norm weighted by $S$.

 For any integer $k\geq 0$, Moving Horizon Estimators are designed to forget about the input and output trajectory before time $k-N$. In this section, we are first interested in the discretized MHE Problem in a batch formulation. Thus one is looking for a state trajectory $(\zeta_{l})_{k-N\leq l\leq k}$ and a vector of landmarks $p$ at the same time that match the visual and inertial measurements. Integrating \eqref{eq:diff_drive} between $t_{k,0}$ and $t_{k+1,0}$ inside an optimisation problem is not computationally tractable. Thus, the state vector $(\zeta_{l})_{k-N\leq l\leq k}$ are linked using \eqref{eq:inertial_meas_1} leading to:      
\begin{align*}
    \zeta_{l+1} =\zeta_{l}+u_l+d_l, \quad l=k-N,\dots,k-1, 
\end{align*}
where $(d_{l})_{k-N\leq l\leq k-1}$ are noise variable added to take into account the presence of disturbations. 
 It is then important to keep track of the knowledge of the past and weigh it in the optimisation problem through an \emph{arrival cost}.   
 Thus, we assume that a state estimate and a landmark estimate at time $k-N$ respectively denoted by $\hat{z}_{k-N}$ and $\hat{\ell}_{k-N}$ are available. We also assume that a weighting matrix denoted by  $\Pi_{k-N}$ is available. Its computation is detailed in section  \ref{sec:arrival_cost_batch}. 
 
 \begin{equation}
\begin{array}{rrclcc}
\displaystyle \underset{\zeta_l, d_l, p}{\text{min}} & \multicolumn{3}{l}{ \left \Vert \begin{bmatrix}\zeta_{k-N}-\hat{z}_{k-N}\\ p-\hat{\ell}_{k-N}\end{bmatrix}\right\Vert_{\Pi_{k-N}^{-1}}^2 +\Vert y_k-A_k h(\zeta_k,p)\Vert^2_{R_{vis}^{-1}}}\vspace{0.05cm}\\ 
 \multicolumn{3}{l}{+\sum_{l=k-N}^{k-1} \Vert d_l\Vert^2_{Q_{in}^{-1}} + \Vert y_l-A_lh(\zeta_l,p)\Vert^2_{R_{vis}^{-1}}}\\
\textrm{s.t.} &\zeta_{l+1} =\zeta_{l}+u_l+d_l, \quad l=k-N,\dots,k-1, 
\end{array}
\tag{$P_{k,ba}$}
\label{prob:MHE_bearing_inertial_visual_disc}
\end{equation}

Note that in Problem \eqref{prob:MHE_bearing_inertial_visual_disc} the resulting dynamics is a discrete time single integrator whose input are the inertial measurements.

 \subsection{Arrival cost computation}\label{sec:arrival_cost_batch}

 The goal of this section is to detail the computation of $\Pi_k$ for any $k\geq0$. First, we fix a matrix $\Pi_0\in \mathbb{S}_{3+2J}^{++}$. Then, for any integer $k\geq0$, $\Pi_k$ is computed using the equation of an Extended Kalman Filter by integrating forward the most recent MHE state and landmark estimate. More precisely,  if we fix some joint estimate $\hat{\xi}_k=(\hat{z}_{k},\hat{\ell}_k)\in \mathbb{R}^{3+2J}$ and a covariance matrix $\Pi_k\in \mathbb{S}_{3+2J}^{++}$ for some $k\geq0$ then for any  $0\leq i \leq m $, we denote by $\xi_{k,i}^+$ the prediction at $t_{k,i}$ from system \eqref{eq:diff_drive_short} with input $\omega_{odo}$. It reads:
\begin{align}
    \xi_{k,i}^+=f_{tot}(\hat{\xi}_{k},\omega_{k,0:i}),
\end{align} 
 where  $\omega_{k,0:i}=(\omega_{k,0},\dots,\omega_{k,i})$ and $f_{tot}(z,\ell,\omega)=\begin{bmatrix}f_{dis,i}(z,\omega)\\ \ell \end{bmatrix}$.
 From this, one can compute the prediction of the covariance matrix up to time $t_{k,i}$ which is denoted by $\Pi_{k,i}^+$. It is defined recursively as follows for any $0\leq i \leq m-1$:
 \begin{align}
     \Pi_{k,0}^+&=\Pi_k,\\
       \Pi_{k,i+1}^+&=F(\xi_{k,i}^+)\Pi_{k,i}F^T(\xi_{k,i}^+)+G(\xi_{k,i}^+)Q^{odo}G^T(\xi_{k,i}^+),
 \end{align}
 where $F(\xi_{k,i}^+)=\nabla_{(z,\ell)}f_{tot}(\hat{\xi}_{k},\omega_{k,0:i})$ and  $G(\xi_{k,i}^+)=\nabla_{\omega}f_{tot}(\hat{\xi}_{k},\omega_{k,0:i})$ and $\nabla f_{tot}$ representing the differential of $f_{tot}$ with respect to the indicated variables.
 The standard correction step at time $t_{k,N}=t_{k+1,0}$ is then applied using the visual measurements \eqref{eq:visual_meas}. We first compute the Kalman gain, which is denoted by $K_{k+1}$ and reads:
 \begin{align}
     K_{k+1}(\xi_{k,m}^+)=&\Pi_{k,m}^+H_{k+1}^T(\xi_{k,m}^+)\\
     &(H_{k+1}(\xi_{k,m}^+)\Pi_{k,m}^+H_{k+1}^T(\xi_{k,m}^+)+R_{vis})^{-1},\notag
 \end{align}
  and leads to the following definition of $\Pi_{k+1}$:
  \begin{align}
      \Pi_{k+1}=\Pi_{k,m}^+-K_{k+1}(\xi_{k,m}^+)H_{k+1}(\xi_{k,m}^+)\Pi_{k,m}^+,
  \end{align}
  where $H_{k+1}(\xi_{k,m}^+)=A_k\nabla_{(z,\ell)} h(\xi_{k,m}^+)$ and $\nabla$ denotes the differential operator.

 \section{Block coordinate Moving horizon estimation for Bearing-only SLAM}\label{sec:bc_MHE}
 The idea of this section is to present the distributed block coordinate version of the problems \eqref{prob:MHE_bearing_inertial_visual_disc} where one looks alternatively for the collection of landmarks for a given state trajectory estimate and for a state trajectory for given landmarks estimates. In order to decouple state and landmark variables the matrices $\Pi_k$ are assumed to block diagonal matrices composed of $J+1$ blocks: one  $3\times 3$ blocks  for state/state correlations only denoted by $\Pi_{zz,k}$, and $J$ $2\times 2 $ blocks for one-by-one landmark/landmark correlations denoted by $(\Pi_{\ell^{(j)}\ell^{(j)},k})_{1\leq j\leq J}$. This assumption implies that the cost in \eqref{prob:MHE_bearing_inertial_visual_disc} is separable with respect to landmarks for a fixed state trajectory estimate which makes distributed resolution possible. 
 \subsection{Distributed landmark estimation for a given state trajectory }
More precisely, let $(\hat{z}_l)_{k-N\leq l \leq k}$ be some estimates of $(z(t_l))_{k-N\leq l \leq k-1}$. By removing constant terms with respect to $p$, the landmark estimation problem reads:
 \begin{equation}
\begin{array}{rrclcc}
\displaystyle \underset{p}{\text{min}} & \multicolumn{3}{l}{ \left \Vert  p-\hat{\ell}_{k-N}\right\Vert_{\Pi_{\ell\ell,k-N}^{-1}}^2  +\sum_{l=k-N}^{k} \Vert y_l-A_lh(\hat{z}_l,p)\Vert^2_{R_{vis}^{-1}}}\vspace{0.5cm} \\
\end{array}
\tag{$P_{k,\ell}$}
\label{prob:MHE_bearing_visual_disc_param_only}
\end{equation}
 Note that \eqref{prob:MHE_bearing_visual_disc_param_only} depends only on the trajectory estimates and not on any dynamics. Besides, if the visual measurement are supposed independent, then $R_{vis}$ is block diagonal with respect to individual landmarks. Since we assumed that $\Pi_{\ell\ell,k-N}$ is block diagonal,  \eqref{prob:MHE_bearing_visual_disc_param_only} can be split and solved landmark by landmark. For any $1\leq j\leq J$, the split problem reads:

 \begin{equation}
\begin{array}{rrclcc}
\displaystyle \underset{p^{(j)}}{\text{min}} & \multicolumn{3}{l}{ \left \Vert  p^{(j)}-\hat{\ell}_{k-N}^{(j)}\right\Vert_{\Pi_{\ell^{(j)}\ell^{(j)},k-N}^{-1}}^2  }\vspace{0.5cm} \\

\multicolumn{3}{l}{+\sum_{l=k-N}^{k} \Vert y_l^{(j)}-a_l^{(j)}h^{(j)}(\hat{z}_l,p^{(j)})\Vert^2_{R_{vis,j}^{-1}}}
\end{array}
\tag{$P_{k,\ell,j}$}
\label{prob:MHE_bearing_visual_disc_1_land}
\end{equation}
where $R_{vis,j}$ is the block of $R_{vis}$ corresponding to landmark $\ell^{j}$.
Consequently if the landmark $j$ is seen at time k (i.e. $a_{k}^{(j)}=1$) then  \eqref{prob:MHE_bearing_visual_disc_1_land} is then solved by a nonlinear programming (NLP) solver using only a fixed number of iterations. 

 \subsection{State estimation for given landmarks estimates}
 In this section, for an integer $k\geq0$, we fix a landmark estimate $\hat{\ell}_k$. 
 Then, the state trajectory estimation subproblem coming from \eqref{prob:MHE_bearing_inertial_visual_disc} reads:  
 \begin{equation}
 \tag{$P_{k,z}$}
\begin{array}{rrclcc}
\displaystyle \underset{\zeta_l, d_l}{\text{min}} & \multicolumn{3}{l}{ \left \Vert\zeta_{k-N}-\hat{z}_{k-N}\right\Vert_{\Pi_{zz,k-N}^{-1}}^2 + \Vert y_k-A_kh(\zeta_k,\hat{\ell}_{k})\Vert^2_{R_{vis}^{-1}} }\vspace{0.05cm} \\
 \multicolumn{3}{l}{+\sum_{l=k-N}^{k-1} \Vert d_l\Vert^2_{Q_{in}^{-1}} + \Vert y_l-A_l h(\zeta_l,\hat{\ell}_{k})\Vert^2_{R_{vis}^{-1}}}\\
\textrm{s.t.} &\zeta_{l+1} =\zeta_{l}+u_l+d_l, \quad l=k-N,\dots,k-1, 
\end{array}
\label{prob:MHE_bearing_inertial_visual_disc_state_only}
\end{equation}
 
This problem can also solved approximately by a NLP solver. Similarly to the batch formulation one obtains an estimate  $\hat{\xi}_k=(\hat{z}_{k},\hat{\ell}_k)\in \mathbb{R}^{3+2J}$.

  \subsection{Arrival cost computation }

 The goal of this section is to detail the computation of a distributed family of the covariance matrices $\Pi_{zz,k}$ and $(\Pi_{\ell{(j)}\ell^{(j)},k})_{j=1,\dots,J}$ for any $k\geq0$ and $\ell=1,\dots,J$. First, we fix a matrices $\Pi_{zz,0}\in\mathbb{S}_{3}^{++} $ and $\Pi_{\ell\ell,0}\in \mathbb{S}_{2}^{++}$ for $\ell=1,\dots,J$. Then, for any integer $k\geq0$, the matrices $\Pi_{zz,k}$ and  $(\Pi_{\ell{(j)}\ell^{(j)},k})_{j=1,\dots,J}$ computed using the equations of an adhoc distributed Extended Kalman Filter. For conciseness, the matrices $\Pi_{\ell^{(j)}\ell^{(j)},k}$ are renamed $\Pi_{jj,k}$ 
 
 \subsubsection{Block Coordinate Prediction step}
  Similarly to the batch version of the EKF from section \ref{sec:arrival_cost_batch}, we fix some joint estimate $\hat{\xi}_k=(\hat{z}_{k},\hat{\ell}_k)\in \mathbb{R}^{3+2J}$ and a covariance matrices $\Pi_{zz,k}\in \mathbb{S}_{3}^{++} $ and  $(\Pi_{jj,k})_{j=1,\dots,J}$ for some $k\geq0$ then for any  $0\leq i \leq m $, we denote by $\xi_{k,i}^+$ the prediction at $t_{k,i}$ from system \eqref{eq:diff_drive_short} with input $\omega_{odo}$. It reads:
\begin{align}
    \xi_{k,i}^+=f_{tot}(\hat{\xi}_{k},\omega_{k,0:i}),
\end{align} 
 where  $\omega_{k,0:i}=(\omega_{k,0},\dots,\omega_{k,i})$ and $f_{tot}(z,\ell,\omega)=\begin{bmatrix}f_{dis,i}(z,\omega)\\ \ell \end{bmatrix}$.
 From this, one can compute the prediction of the covariance matrices up to time $t_{k,i}$ which are denoted by $\Pi_{zz,k,i}^+\in \mathbb{S}_{3}^{++} $ and  $(\Pi_{jj,k,i}^+)_{j=1,\dots,J}$. It is defined recursively as follows for any $0\leq i \leq m$, and any $j=1,\dots,J$:
 \begin{align}
     \Pi_{zz,k,0}^+&=\Pi_{zz,k},\\
       \Pi_{jj,k,0}^+&=\Pi_{jj,k},\\
       \Pi_{zz,k,i+1}^+&=F(\xi_{k,i}^+)\Pi_{zz,k,i}F^T(\xi_{k,i}^+)+G(\xi_{k,i}^+)Q^{odo}G^T(\xi_{k,i}^+),\\
       \Pi_{jj,k,i+1}^+&=\Pi_{jj,k,i}^+
 \end{align}
 where $F(\xi_{k,i}^+)=\nabla_{(z,\ell)}f_{dis,i}(\hat{\xi}_{k},\omega_{k,0:i})$ and  $F(\xi_{k,i}^+)=\nabla_{\omega}f_{dis,i}(\hat{\xi}_{k},\omega_{k,0:i})$ and $\nabla f_{dis,i}$ representing the differential of $f_{tot}$ with respect to the indicated variables.
\subsubsection{Distributed Block Coordinate Correction step}

First, a block coordinate correction step at time $t_{k,N}=t_{k+1,0}$ is then applied using the visual measurements \eqref{eq:visual_meas}.
We first make an approximation of the visual observation equation:
\begin{align*}
y_{k+1}\approx A_{k+1}(h(\hat{z}_k^+ +w_k^+,\ell)+v_{k+1})    
\end{align*}
where $w_k \sim \mathcal{N}(0, \Pi_{zz,k}^+)$. Moreover, landmark $j$ is updated at time $k$ only if it is seen at that time i.e. when $a_k^{(j)}=1$. We assume $w_k^+$ and $v_{k+1}$ are independent and we compute the Kalman gain for each landmark which  denoted by $K_{j,k+1}$ and reads for any $j=1,\dots,J$, if $a_k^{(j)}=1$ then:
 \begin{align}
     &K_{j,k+1}(\xi_{k,m}^+)=\Pi_{j,k}^+H_{j,k+1}^T(\xi_{k,m}^+)\label{eq:dist_param_update_1}\\
     &[H_{j ,k+1}(\xi_{k,m}^+)\Pi_{jk}^+H_{j,k+1}^T(\xi_{k,m}^+)\\
     &+R_{vis}+H_{z,k+1}(\xi_{k,m}^+)\Pi_{zz,k}H^T_{z,k+1}(\xi_{k,m}^+)]^{-1},
 \end{align}
  which leads to the following definition of $\Pi_{jj,k+1}$:
  \begin{align}
      \Pi_{jj,k+1}=\Pi_{jj,k}-K_{j,k+1}(\xi_{k,m}^+)H_{j,k+1}(\xi_{k,m}^+)\Pi_{jj,k},\label{eq:dist_param_update_2}
  \end{align}
  where $H_{j,k+1}(\xi_{k,m}^+)=a_{k+1}^{(j)}\nabla_{\ell^{(j)}}h^{(j)}(\xi_{k,m}^+)$.
  Otherwise if $a_k^{(j)}=0$ then $\Pi_{jj,k+1}=\Pi_{jj,k}$.
 
  One now assumes  that some  updated  estimate of the collection of landmarks $\hat{\ell}_{k+1}$ has been computed. We make the following approximation of the visual observation equation
  \begin{align*}
y_{k+1}\approx A_{k+1}(h(z_k,\hat{\ell}_{k+1}+w_{\ell,k+1})+v_{k+1})    
\end{align*}
where $w_{\ell,k+1} \sim \mathcal{N}(0, \Pi_{\ell\ell,k}^+)$. Then, we assume $w_{\ell,k+1}$ and $v_{k+1}$ are independent and we compute the Kalman gain for the robot's state which  denoted by $K_{z,k+1}$ and reads:

 \begin{align}
     &K_{z,k+1}(\xi_{k,m}^+)=\Pi_{zz,k,m}^+H_{z,k+1}^T(\xi_{k,m}^+)\\
     &(H_{z ,k+1}(\xi_{k,m}^+)\Pi_{zz,k,m}^+H_{j,k+1}^T(\xi_{k,m}^+)\\
     &+R_{vis}+H_{\ell,k+1}\Pi_{\ell\ell,k+1}H^T_{\ell,k+1})^{-1},
 \end{align}
  which leads to the following definition of $\Pi_{jj,k+1}$:
  \begin{align}
      \Pi_{zz,k+1}=\Pi_{zz,k}-K_{z,k+1}(\xi_{k,m}^+)H_{z,k+1}(\xi_{k,m}^+)\Pi_{zz,k,m}^+,
  \end{align}
  where $H_{\ell,k+1}(\xi_{k,m}^+)=A_{k+1}\nabla_{\ell} h(\xi_{k,m}^+)$.

 The assumption that the matrices $\Pi_{k,\ell\ell}$ are block diagonal is strong because it means that one neglects the correlations between landmarks. However, the proposed distributed Kalman covariance update allows one to reintroduce correlation between the landmark and state estimates which seems to be enough to get a good confidence measure on the state and landmarks to be used as an arrival cost in the MHE problems. 
\section{Algorithm} \label{sec:algo}
  \begin{algorithm}[h]
{\setlength{\baselineskip}{1\baselineskip}
\caption{Block coordinate Descent in MHE for SLAM}\label{algo_MHE_dist}
\begin{algorithmic}[1]
\State Fix $\hat{z}_0=z_0$ and $\xi_0=z_0$ 
\State Choose $\hat{\ell}_0$ and $\Pi_0$
\State Get $y_0$

\For{$k=1,2,\dots$}

\State Get $y_k$.
\State Compute $\Pi_{k}$.
\If{$k<N$}

\State Get $\hat{z}_k$ by integrating \eqref{eq:diff_drive_short} from $t_{k-1}$ to $t_k$ with input $\omega_{odo}$.
\For {$j=1,\dots,J$}
\If{$a_k^{(j)}=1$}
\State Solve \eqref{prob:MHE_bearing_visual_disc_1_land} at $(\hat{z}_0,\dots,\hat{z}_k)$ by setting $N=k+1$ and get an optimal solution $p^{(j)*}$.
\State  $\hat{\ell}_k^{(j)}\leftarrow p^{(j)*}$.
\EndIf
\EndFor
\Else 
\State Compute $\hat{z}^{+}_k$ using \eqref{eq:diff_drive_short} with input $\omega_{odo}$.

\For {$j=1,\dots,J$}
\If{$a_k^{(j)}=1$}
\State Solve \eqref{prob:MHE_bearing_visual_disc_1_land}at $(\hat{z}_{k-N},\dots,\hat{z}_{k-1},z^{+}_{k-1})$ and get an optimal solution $p^{(j)*}$.
\State  $\hat{\ell}_k^{(j)}\leftarrow p^{(j)*}$.
\EndIf
\EndFor
\State Solve \eqref{prob:MHE_bearing_inertial_visual_disc_state_only}  at $\hat{\ell}_k$ and get  $\zeta_{k-N}^*,\dots,\zeta_k^*$.
\State $\hat{z}_k\leftarrow \zeta_k^*$.
\EndIf
\State 
\EndFor
\end{algorithmic}
\par}
\end{algorithm}
  The resulting algorithm is summarised in Algorithm \ref{algo_MHE_dist}.
\begin{remark}\hfill

\begin{itemize}

    \item The main advantage of  Algorithm \ref{algo_MHE_dist} is that the landmark update can be parallelised  since both the MHE problem \eqref{prob:MHE_bearing_visual_disc_1_land} and the Kalman update for the arrival cost \eqref{eq:dist_param_update_1}-\eqref{eq:dist_param_update_2} are distributed over the landmarks variables.
    \item Since the landmarks that are not seen at time $k$ are not updated, the classical effect of loop closure that allows the filter to correct every landmark at the same time cannot happen. However, as shown in Section \ref{sec:simulations}, several loops ensures that all the maps is properly updated. 
    
 
\end{itemize}
 
\end{remark}
A key factor for good performance of such Moving Horizon Estimation schemes are observability conditions, see \cite{flayac_non-uniform_nodate, knufer_time-discounted_2020}.
Because of the nonlinearities in the measurement and dynamic equations, observability properties depend on the trajectory of the extended system state/landmark and so might the estimation error. Thus, the goal of Section \ref{sec:simulations} is to illustrate the results of the proposed estimation algorithm for different robot trajectories and different environments. 
\section{Simulations}\label{sec:simulations}

\begin{figure}

     \centering
     \begin{subfigure}[b]{0.49\textwidth}
         \centering
        \includegraphics[width=\textwidth]{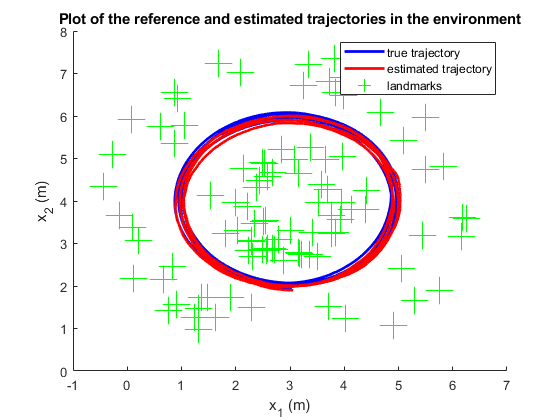}
        \caption{}
        \label{fig:est_circle}
     \end{subfigure}
     
     \hfill
     \begin{subfigure}[b]{0.49\textwidth}
         \centering
         \includegraphics[width=\textwidth]{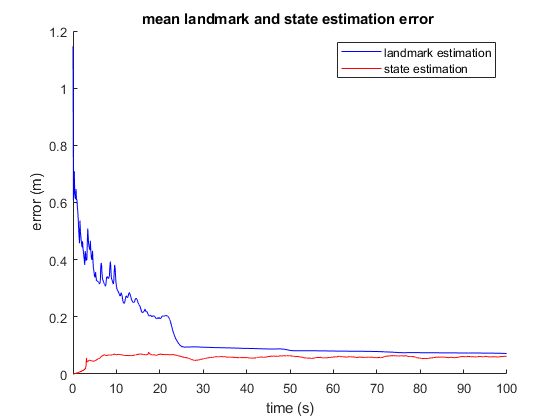}
         \caption{}
         \label{fig:error_circle}
     \end{subfigure}
     \caption{Plots of a sample circular trajectory, its estimation and the corresponding average landmark and state estimation errors for $50$ Monte-Carlo simulations.}
\end{figure}

In this section, we present simulations of a 2D environments with with $2$ configurations of $J=100$ landmarks and $3$ types of trajectories, a circular one, a straight one and one with wiggles. Figure \ref{fig:est_circle} shows a example of circular trajectory with several loops with landmarks dispatched in a double ring. The parameters of the robot from \eqref{eq:diff_drive} have been chosen as characteristics of a standard Jackal robot knowing: $D=0.043$, $R_r =R_l= 0.1$. The noise covariance have been set as follows: $R_{vis}=0.001I$, $Q_{in}=10^{-4}\Delta_{vis}I_3$, and $Q_{odo}=0.009I_2$, where $I$ denotes the identity matrix of appropriate dimension. The parameters $a_k^{(j)}$, representing data association are assumed to be given without error, for any $1\leq j\leq J$ and any $k\geq0$.  Besides, a maximal range has been implemented  on the bearing sensor through the variables $a_k^{(j)}$. It is of $2m$ for the circular scenarios and $3.6m$ for the two others. Running times have not been included since code optimisation is not the topic of this paper and the actual parallelisation process of the distributed scheme has not been implemented yet.    

\begin{figure}
     \centering
     \begin{subfigure}[b]{0.49\textwidth}
         \centering
        \includegraphics[width=\textwidth]{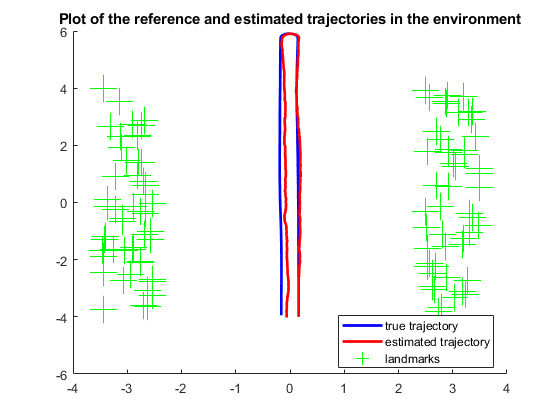}
        \caption{}
         \label{fig:traj_straight_line}
     \end{subfigure}
     \hfill
     \begin{subfigure}[b]{0.49\textwidth}
         \centering
         \includegraphics[width=\textwidth]{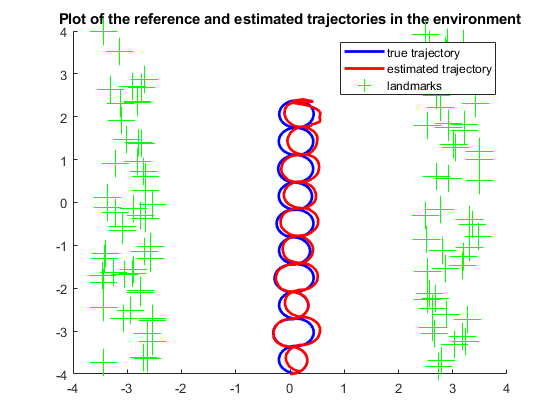}
         \caption{}
         \label{fig:traj_snaking}
     \end{subfigure}
     \caption{Plots of a sample straight and a wiggling trajectory with their estimates in the same environment}
\end{figure}

\begin{figure}
     \centering
     \begin{subfigure}[b]{0.49\textwidth}
         \centering
        \includegraphics[width=\textwidth]{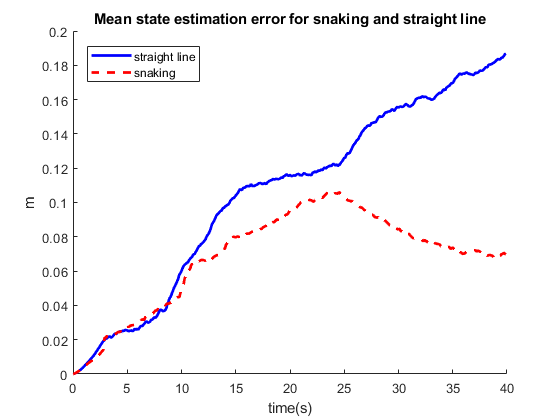}
 \caption{}
         \label{fig:error_state}
     \end{subfigure}
     \hfill
     \begin{subfigure}[b]{0.49\textwidth}
         \centering
         \includegraphics[width=\textwidth]{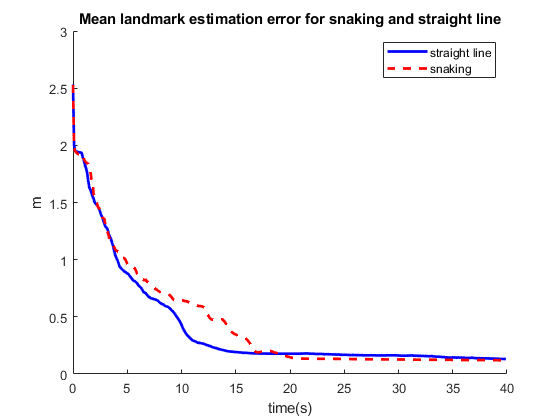}
         \caption{}
         \label{fig:error_land}
     \end{subfigure}
     \caption{Plots of the average state and landmark errors in the case of a straight and wiggling trajectory in the same environment after $50$ Monte-Carlo simulations.}
\end{figure}

The performance of the proposed method in this case is demonstrated in Figure \ref{fig:error_circle} where both the state and mean landmark estimation error are converging to a small value. Note that the initial state estimation error is assumed to be zero since the initial position and orientation of the robot is assumed to be known.  Observability theory coming from circumnavigation   \cite{deghat_target_2010, deghat_localization_2014, de_marco_position_2020, fossen_observers_2017, johansen_globally_nodate,flayac_non-uniform_nodate} suggest that circular patterns should improve estimation performance. Figure \ref{fig:traj_straight_line} and \ref{fig:traj_snaking} show an example of a back-and-forth straight and snaking trajectory in a corridor-like environment with only landmarks on the side. Figure \ref{fig:error_state} and \ref{fig:error_land} show that, as expected, the wiggling patterns are helping the estimation process which result in a smaller state estimation error than in the case of a straight trajectory. 

\section{Conclusion}

In this paper, a block coordinated Moving Horizon Estimation algorithm for Visual Inertial Odometry SLAM is proposed. It is leveraging ideas coming from Bundle Adjustement, nonlinear estimation and nonlinear programming in order to make updates of the landmarks distributed. The performance of the proposed method is demonstrated in the presence of noise is demonstrated through simulations. 

\bibliographystyle{plain}
\bibliography{bibfile}

\begin{thebibliography}{10}

\bibitem{bertsekas_nonlinear_1997}
D.~P. Bertsekas.
\newblock Nonlinear {Programming}.
\newblock {\em Journal of the Operational Research Society}, 48(3):334--334,
  1997.
\newblock mlzsync1:0048\{"extrafields":\{"publisher":"Taylor \&
  Francis"\}\}\_eprint: https://doi.org/10.1057/palgrave.jors.2600425.

\bibitem{bustos_visual_2019}
Alvaro~Parra Bustos, Tat-Jun Chin, Anders Eriksson, and Ian Reid.
\newblock Visual {SLAM}: {Why} {Bundle} {Adjust}?
\newblock In {\em 2019 {International} {Conference} on {Robotics} and
  {Automation} ({ICRA})}, pages 2385--2391, Montreal, QC, Canada, May 2019.
  IEEE.

\bibitem{campos_orb-slam3_2021}
Carlos Campos, Richard Elvira, Juan J.~Gomez Rodriguez, Jose~M.M Montiel, and
  Juan~D. Tardos.
\newblock {ORB}-{SLAM3}: {An} {Accurate} {Open}-{Source} {Library} for
  {Visual}, {Visual}–{Inertial}, and {Multimap} {SLAM}.
\newblock {\em IEEE Transactions on Robotics}, 37(6):1874--1890, December 2021.

\bibitem{de_marco_position_2020}
S.~de~Marco, M-D. Hua, T.~Hamel, and C.~Samson.
\newblock Position, {Velocity}, {Attitude} and {Accelerometer}-{Bias}
  {Estimation} from {IMU} and {Bearing} {Measurements}.
\newblock In {\em 2020 {European} {Control} {Conference} ({ECC})}, pages
  1003--1008, Saint Petersburg, Russia, May 2020. IEEE.

\bibitem{deghat_target_2010}
Mohammad Deghat, Iman Shames, Brian D.~O. Anderson, and Changbin Yu.
\newblock Target localization and circumnavigation using bearing measurements
  in {2D}.
\newblock In {\em 49th {IEEE} {Conference} on {Decision} and {Control}
  ({CDC})}, pages 334--339, Atlanta, GA, USA, December 2010. IEEE.

\bibitem{deghat_localization_2014}
Mohammad Deghat, Iman Shames, Brian D.~O. Anderson, and Changbin Yu.
\newblock Localization and {Circumnavigation} of a {Slowly} {Moving} {Target}
  {Using} {Bearing} {Measurements}.
\newblock {\em IEEE Transactions on Automatic Control}, 59(8):2182--2188,
  August 2014.

\bibitem{flayac_non-uniform_nodate}
Emilien Flayac and Iman Shames.
\newblock Non-uniform {Observability} for {Fast} {Moving} {Horizon}
  {Estimation} with application to the {SLAM} problem.
\newblock page~19.

\bibitem{fuentes-pacheco_visual_2015}
Jorge Fuentes-Pacheco, José Ruiz-Ascencio, and Juan~Manuel Rendón-Mancha.
\newblock Visual simultaneous localization and mapping: a survey.
\newblock {\em Artificial Intelligence Review}, 43(1):55--81, January 2015.

\bibitem{heo_ekf-based_2019}
Sejong Heo, Jaehyuck Cha, and Chan~Gook Park.
\newblock {EKF}-{Based} {Visual} {Inertial} {Navigation} {Using} {Sliding}
  {Window} {Nonlinear} {Optimization}.
\newblock {\em IEEE Transactions on Intelligent Transportation Systems},
  20(7):2470--2479, July 2019.

\bibitem{bebis_monocular_2016}
Timo Hinzmann, Thomas Schneider, Marcin Dymczyk, Andreas Schaffner, Simon
  Lynen, Roland Siegwart, and Igor Gilitschenski.
\newblock Monocular {Visual}-{Inertial} {SLAM} for {Fixed}-{Wing} {UAVs}
  {Using} {Sliding} {Window} {Based} {Nonlinear} {Optimization}.
\newblock In George Bebis, Richard Boyle, Bahram Parvin, Darko Koracin, Fatih
  Porikli, Sandra Skaff, Alireza Entezari, Jianyuan Min, Daisuke Iwai, Amela
  Sadagic, Carlos Scheidegger, and Tobias Isenberg, editors, {\em Advances in
  {Visual} {Computing}}, volume 10072, pages 569--581. Springer International
  Publishing, Cham, 2016.
\newblock Series Title: Lecture Notes in Computer Science.

\bibitem{huang_review_2019}
Shoudong Huang.
\newblock A review of optimisation strategies used in simultaneous localisation
  and mapping.
\newblock {\em Journal of Control and Decision}, 6(1):61--74, January 2019.

\bibitem{huang_critique_2016}
Shoudong Huang and Gamini Dissanayake.
\newblock A critique of current developments in simultaneous localization and
  mapping.
\newblock {\em International Journal of Advanced Robotic Systems},
  13(5):172988141666948, September 2016.

\bibitem{ila_information-based_2010}
V.~Ila, J.M. Porta, and J.~Andrade-Cetto.
\newblock Information-{Based} {Compact} {Pose} {SLAM}.
\newblock {\em IEEE Transactions on Robotics}, 26(1):78--93, February 2010.

\bibitem{jia_survey_2019}
Yujiao Jia, Xinying Yan, and Yihan Xu.
\newblock A {Survey} of simultaneous localization and mapping for robot.
\newblock In {\em 2019 {IEEE} 4th {Advanced} {Information} {Technology},
  {Electronic} and {Automation} {Control} {Conference} ({IAEAC})}, pages
  857--861, Chengdu, China, December 2019. IEEE.

\bibitem{johansen_globally_nodate}
Tor~A Johansen and Edmund Brekke.
\newblock Globally {Exponentially} {Stable} {Kalman} {Filtering} for {SLAM}
  with {AHRS}.
\newblock page~8.

\bibitem{karrer_cvi-slamcollaborative_2018}
Marco Karrer, Patrik Schmuck, and Margarita Chli.
\newblock {CVI}-{SLAM}—{Collaborative} {Visual}-{Inertial} {SLAM}.
\newblock {\em IEEE Robotics and Automation Letters}, 3(4):2762--2769, October
  2018.

\bibitem{kishimoto_moving_2019}
Yosuke Kishimoto, Kiyotsugu Takaba, and Asuka Ohashi.
\newblock Moving {Horizon} {Multi}-{Robot} {SLAM} {Based} on {C}/{GMRES}
  {Method}.
\newblock In {\em 2019 {International} {Conference} on {Advanced} {Mechatronic}
  {Systems} ({ICAMechS})}, pages 22--27, Kusatsu, Shiga, Japan, August 2019.
  IEEE.

\bibitem{klein_parallel_2007}
Georg Klein and David Murray.
\newblock Parallel {Tracking} and {Mapping} for {Small} {AR} {Workspaces}.
\newblock In {\em 2007 6th {IEEE} and {ACM} {International} {Symposium} on
  {Mixed} and {Augmented} {Reality}}, pages 1--10, Nara, Japan, November 2007.
  IEEE.

\bibitem{knufer_time-discounted_2020}
Sven Knufer and Matthias~A. Muller.
\newblock Time-{Discounted} {Incremental} {Input}/{Output}-to-{State}
  {Stability}.
\newblock In {\em 2020 59th {IEEE} {Conference} on {Decision} and {Control}
  ({CDC})}, pages 5394--5400, Jeju, Korea (South), December 2020. IEEE.

\bibitem{fossen_observers_2017}
Florent Le~Bras, Tarek Hamel, Robert Mahony, and Claude Samson.
\newblock Observers for {Position} {Estimation} {Using} {Bearing} and {Biased}
  {Velocity} {Information}.
\newblock In Thor~I. Fossen, Kristin~Y. Pettersen, and Henk Nijmeijer, editors,
  {\em Sensing and {Control} for {Autonomous} {Vehicles}}, volume 474, pages
  3--23. Springer International Publishing, Cham, 2017.
\newblock Series Title: Lecture Notes in Control and Information Sciences.

\bibitem{liu_ice-ba_2018}
Haomin Liu, Mingyu Chen, Guofeng Zhang, Hujun Bao, and Yingze Bao.
\newblock {ICE}-{BA}: {Incremental}, {Consistent} and {Efficient} {Bundle}
  {Adjustment} for {Visual}-{Inertial} {SLAM}.
\newblock In {\em 2018 {IEEE}/{CVF} {Conference} on {Computer} {Vision} and
  {Pattern} {Recognition}}, pages 1974--1982, Salt Lake City, UT, June 2018.
  IEEE.

\bibitem{lupton_visual-inertial-aided_2012}
Todd Lupton and Salah Sukkarieh.
\newblock Visual-{Inertial}-{Aided} {Navigation} for {High}-{Dynamic} {Motion}
  in {Built} {Environments} {Without} {Initial} {Conditions}.
\newblock {\em IEEE Transactions on Robotics}, 28(1):61--76, February 2012.

\bibitem{mur-artal_visual-inertial_2017}
Raul Mur-Artal and Juan~D. Tardos.
\newblock Visual-{Inertial} {Monocular} {SLAM} with {Map} {Reuse}.
\newblock {\em IEEE Robotics and Automation Letters}, 2(2):796--803, April
  2017.
\newblock arXiv: 1610.05949.

\bibitem{strasdat_visual_2012}
Hauke Strasdat, J.M.M. Montiel, and Andrew~J. Davison.
\newblock Visual {SLAM}: {Why} filter?
\newblock {\em Image and Vision Computing}, 30(2):65--77, February 2012.

\bibitem{taketomi_visual_2017}
Takafumi Taketomi, Hideaki Uchiyama, and Sei Ikeda.
\newblock Visual {SLAM} algorithms: a survey from 2010 to 2016.
\newblock {\em IPSJ Transactions on Computer Vision and Applications}, 9(1),
  December 2017.

\bibitem{thrun_probabilistic_2002}
Sebastian Thrun.
\newblock Probabilistic robotics.
\newblock {\em Communications of the ACM}, 45(3):52--57, March 2002.

\bibitem{thrun_graph_2006}
Sebastian Thrun and Michael Montemerlo.
\newblock The {Graph} {SLAM} {Algorithm} with {Applications} to {Large}-{Scale}
  {Mapping} of {Urban} {Structures}.
\newblock {\em The International Journal of Robotics Research},
  25(5-6):403--429, May 2006.

\bibitem{valencia_mapping_2018}
Rafael Valencia and Juan Andrade-Cetto.
\newblock {\em Mapping, {Planning} and {Exploration} with {Pose} {SLAM}},
  volume 119 of {\em Springer {Tracts} in {Advanced} {Robotics}}.
\newblock Springer International Publishing, Cham, 2018.

\end{thebibliography}

\end{document}